\documentclass[letterpaper, 10 pt, conference]{ieeeconf}  % Comment this line out if you need a4paper

\IEEEoverridecommandlockouts                              % This command is only needed if 
\usepackage{graphicx}
\usepackage[hidelinks]{hyperref}
\title{\LARGE \bf
Trust and Acceptance of Multi-Robot Systems ``in the Wild”. A Roadmap exemplified within the EU-Project BugWright2
}

\author{Pete Schroepfer$^{1,3}$, Nathalie Schauffel$^{4}$, Jan Gr\"undling$^{4}$, Thomas Ellwart$^{4}$, Benjamin Weyers$^{4}$, C\'edric Pradalier$^{1,2}$%} 
\thanks{*This project has received funding from the European Union’s Horizon 2020 research and innovation programme under Grant Agreement No. 871260. Corresponding author: {\tt \footnotesize cedric.pradalier@georgiatech-metz.fr}}
\thanks{$^{1}$CNRS IRL2958 GT-CNRS, Metz, France}%
\thanks{$^{2}$GeorgiaTech Lorraine, Metz, France}%
\thanks{$^{3}$Georgia Institute of Technology, Atlanta, USA}%
\thanks{$^{4}$Trier University, Trier, Germany}%
}

\begin{document}

\maketitle
\thispagestyle{empty}
\pagestyle{empty}

%%%%%%%%%%%%%%%%%%%%%%%%%%%%%%%%%%%%%%%%%%%%%%%%%%%%%%%%%%%%%%%%%%%%%%%%%%%%%%%%
%%%%%%%%%%%%%%%%%%%%%%%%%%%%%%%%%%%%%%%%%%%%%%%%%%%%%%%%%%%%%%%%%%%%%%%%%%%%%%%%
\begin{abstract}
    This paper outlines a roadmap to effectively leverage shared mental models in multi-robot, multi-stakeholder scenarios, drawing on experiences from the BugWright2 project. The discussion centers on an autonomous multi-robot systems designed for ship inspection and maintenance. A significant challenge in the development and implementation of this system is the calibration of trust. To address this, the paper proposes that trust calibration can be managed and optimized through the creation and continual updating of shared and accurate mental models of the robots. Strategies to promote these mental models, including cross-training, briefings, debriefings, and task-specific elaboration and visualization, are examined. Additionally, the crucial role of an adaptable, distributed, and well-structured user interface (UI) is discussed.
\end{abstract}

\section{INTRODUCTION} \label{intro}
BugWright2 is a research project developing an autonomous multi-robot system for inspection and maintenance of medium to large ships through a collaboration of multiple partners in academia and industry. These partners include roboticists, psychologists, user interface (UI) developers, and industrial maritime companies. The system employs Micro Aerial Vehicles (MAVs), Autonomous Underwater Vehicles (AUVs), and magnetic-wheeled crawlers, aiming to streamline maritime operations and reduce reliance on labor-intensive manual processes. One key goal of this project is to enhance worker safety by automating potentially hazardous tasks. At the same time, calibrating trust in Human-Robot Interaction (HRI), particularly in the context of autonomous multi-robot systems like BugWright2, presents significant challenges. Here, each stakeholder’s level of trust needs to be calibrated in to avoid under-usage (not enough trust) and over-reliance (too much trust).

Such trust calibration becomes increasingly complex when applied to real-world scenarios involving multiple "trustors" and "trustees" interacting within different specific contexts. With multiple trustors involved - including operators, supervisors, safety inspectors, and other stakeholders - the trust dynamics become multi-layered in addition to being multidimensional \cite{Pastra_2022}. At the outset, different trustors have different levels of trust for the same trustees, are influenced by their role, their experiences, their understanding of the technology, and individual dispositions (cf. multiple layers of trust in \cite{hoff_trust_2015}). Barring any planning and specific design choices, over time, trust levels will also likely further vary, as trust might increase and decreases differently for different individuals depending on situational experiences and HRI. It is our position, that one of the best ways to calibrate and manage trust in complex real-world environments is by ensuring accurate and shared mental models of the robots (e.g., the mental representations of the knowledge of system features, functionalities, maintenance, and sufficient use). Below, we begin by exploring the multifaceted trust issues in multi-robot/multi-stakeholder settings, share context-specific lessons learned, and finally, discuss how human-robot interaction and human-computer interaction principles aid in UI design to tackle these challenges.

\section{Trust Problem Space for Multi-Robot/Multi-Stakeholder Situations} When viewed at the individual level, trust in HRI can be understood as a complex system consisting of several components and relationships \cite{Schroepfer2023}. Among the trust key components are the trustor, the trustee, and the context in which they reside. In addition, there are numerous internal and external factors that affect the trustor’s internal state and behavior, both at the outset and over time, such as the design of the robots, the trustor’s prior experience, and the trustor’s observations of the robots performing the task (e.g., \cite{ellwart2023human, hancock_evolving_2021, hoff_trust_2015, Pastra_2022, schaefer2016meta}). In a multi-stakeholder/multi-robot scenario, this trust dynamic then exists not only between the individual stakeholders but each stakeholder and each robot.

While this particular scenario adds to the complexity of the trust dynamic, it is possible to leverage the complex environment in a way that is not possible with one-to-one HRI through shared mental models. In the one-on-one case, typically the human participant will start with an initial mental model of the robot based on prior experience and first impressions. Over time, this model is only really updated through the interaction with and observation of the robot (cf., \cite{hoff_trust_2015}). By contrast, in a multi-stakeholder/multi-robot scenario, mental models can be updated by observations and interactions with both other stakeholders and other robots. For example, an observer watching a robot move in a sort of erratic manner might normally view this as a defect by some stakeholders. Yet here, this individual could also look at the operator and, if the operator appeared to be in a normal state, might then assume the behavior is normal. Similarly, when observing one robot, say the magnetic-wheeled crawler, operating in conjunction with a second robot, say a MAV. An observer may update their mental model of the magnetic-wheeled crawler based on observations of the MAV, as the MAV's presence may lead to inferences about the crawlers capabilities and state. While this scenario increases the factors affecting the mental model, it also becomes possible to design both the scenario and the other robots in a way to generate a cohesive, accurate shared mental model between the stakeholders. 

\section{Shared Mental Models: A Roadmap}
\subsection{Cross-Training} In a team situation, a common way of creating a shared mental model is through cross-training, whereby team members use various techniques to share information between members. In HRI, and particularly in the BugWright2 project, this method has been effective when working with members of different backgrounds. For example, prior to an inspection, the roboticists who built and designed the robot could explain each robot’s particular abilities as roles in a scenario. By sharing this information, the other participants can then absorb this data into their mental models, creating greater consistency in the expectations of the robot. On the side of Human Resource (HR) tools here, guidelines on how to conduct a guided reflection on knowledge needs and elements stimulate cross-training and interdisciplinary exchange. 

\subsection{Briefing and Debriefing}Like cross-training, in the BugWright2 project briefing and debriefing have been a very effective means of building and exploiting a shared mental model. As noted above, the participants in a multi-robot multi-stakeholder task often come from a variety of backgrounds and experiences that lead to a large initial variance regarding the mental models of each of the robots. By briefing and debriefing at the outset and scheduled points during the larger task, these mentally disparate mental models are all brought to the forefront and in some ways realigned. This creates a strong base, but then experience creates divergence over time. Continued debriefings, however, provide a correction and adjustment for this divergence. For this project in particular, this has led to greater cohesion and more calibrated trust levels. 

\subsection{Elaboration and Visualization} As a third element to foster accurate and shared mental models within BugWright2 and thus guarantee a well-calibrated level of trust, the method \textit{Task-Specific Elaboration and Visualization} (TSE+V) of work tasks (e.g., steel plate thickness measurement) was applied. Integrating methods of psychological work analysis (e.g., \cite{GROTE2000267}) and business process modeling (e.g., BPMN) TSE+V helps to augment team cognition (e.g., perception of interdependencies, non-routine situations, system features, roles, and responsibilities). Similarities and differences in perceptions and evaluations can thus be mapped, synchronized, and visualized, resulting in mental models that are accurate and shared among different stakeholders, including end-users. Misperceptions can be detected. Critical reflection can prevent both over- and undertrust.

\section{UI Design for Shared Mental Models}
Another important design factor for multi-robot systems is the use of UI. Concerning shared mental models, the design of the UI can heavily influence the amount, format, and quantity of feedback provided by the robot system to each stakeholder. A shared mental model represents the shared understanding of both the functionality of each team member and a sense of the current state. Therefore, designing the UI such that it distributes information evenly and in a way that respects the needs of each stakeholder, can create a greater level of coherence in the mental model \cite{grundling2023example}. In the BugWright2 project, we have found it affecting to provide varying UI options both in terms of medium (screen, virtual reality (VR), augmented reality (AR), etc.) and in terms of content (role-based UI options). For example, by developing a web-based visualization system, stakeholders can observe the state of all the robots and request state information from any device connected to the internet or local network. This not only ensures that all stakeholders receive similar information simultaneously, but that each stakeholder may interactively learn more about the current state of the task and system. Furthermore, by allowing each user to customize their particular means of viewing the information, this also allows different users to view the information in a manner more suited for their consumption. The net result is that stakeholders with varying backgrounds can all get the most current state in a way best suited to them, thereby enabling greater coherence in their mental models.

This is further extended by developing the system in a way that allows both AR and VR mediums. AR and VR through a distributed web system allow stakeholders to observe the state of the task and robots spatially. While this may be particularly helpful for operators, it also may allow any stakeholder that prefers a more immersive perspective to access the current information. Further, embedding questionnaires into VR has been shown to be a valuable method to evaluate subjective experiences during its use \cite{grundling2022answering}. This way, the calibration of trust and shared mental models can be evaluated seamlessly during use.

\section{Conclusion}
Multi-robot multi-stakeholder situations provide unique and interesting challenges for calibrating trust. While these situations may increase the total complexity of the trust dynamic, leveraging concepts and methodologies from research on shared mental models in teams can not only address much of the added complexity but potentially aid in trust calibration. This can, in turn, be enhanced by creating a UI that can be distributed easily to multiple stakeholders.

\bibliographystyle{IEEEtran}
\bibliography{scrita}

\end{document}